\documentclass[10pt,twocolumn,letterpaper]{article}

\usepackage[pagenumbers]{cvpr} 

\definecolor{cvprblue}{rgb}{0.21,0.49,0.74}
\usepackage[pagebackref,breaklinks,colorlinks,allcolors=cvprblue]{hyperref}
\usepackage{multirow}
\usepackage{colortbl}
\usepackage{makecell}

\usepackage{listings}
\usepackage{xcolor}
\def\confName{CVPR}
\def\confYear{2026}

\title{Depth Hypothesis Guided Iterative Refinement for Event–Image Monocular Depth Estimation}

\author{{Daikun Liu}\qquad{Teng Wang\thanks{Corresponding author:~Teng Wang.}}\qquad{Changyin Sun}\\
Southeast University\\
{\tt\small \{dkliu, wangteng, cysun\}@seu.edu.cn}}

\begin{document}
\maketitle
\begin{abstract}
Event cameras hold excellent dynamic properties, showing great potential for monocular depth estimation (MDE). However, existing methods mainly improve performance by optimizing contextual features, but still struggle with the ill-posed and nonlinear nature of direct full-depth regression. In this paper, we propose HypoDepth, the first event–image monocular depth iterative refinement framework. By introducing a discrete Depth Hypothesis Volume (DHV), we transform the depth regression problem into a constrained depth search task. Specifically, we construct a 3D cost volume between the DHV features and contextual features and perform a multi-scale correlation search to guide stable residual optimization. This lightweight cost volume enables efficient global-to-local refinement across multi-resolution. Our method outperforms existing approaches on DSEC and MVSEC with state-of-the-art results and strong zero-shot generalization. Meanwhile, our tiny model achieves an excellent balance between accuracy and efficiency, enabling real-time performance on resource-limited devices.


\end{abstract}    
\section{Introduction}
\label{sec:intro}

Monocular depth estimation (MDE) serves as a foundational task for 3D scene understanding, yet image-based approaches~\cite{DPT, zhang2022hierarchical, ranftl2020towards} often struggle under challenging conditions such as high-speed motion or poor illumination. Event cameras~\cite{survey, dvs}, with their high temporal resolution and dynamic range, can capture meaningful scene changes in such scenarios. However, they lack visual information in static regions, making the recovery of static structures difficult. Consequently, leveraging the complementarity of these two modalities holds great promise for robust MDE.

\begin{figure}[t]
\centering
\includegraphics[width=1.0\linewidth, height=0.21\textwidth]{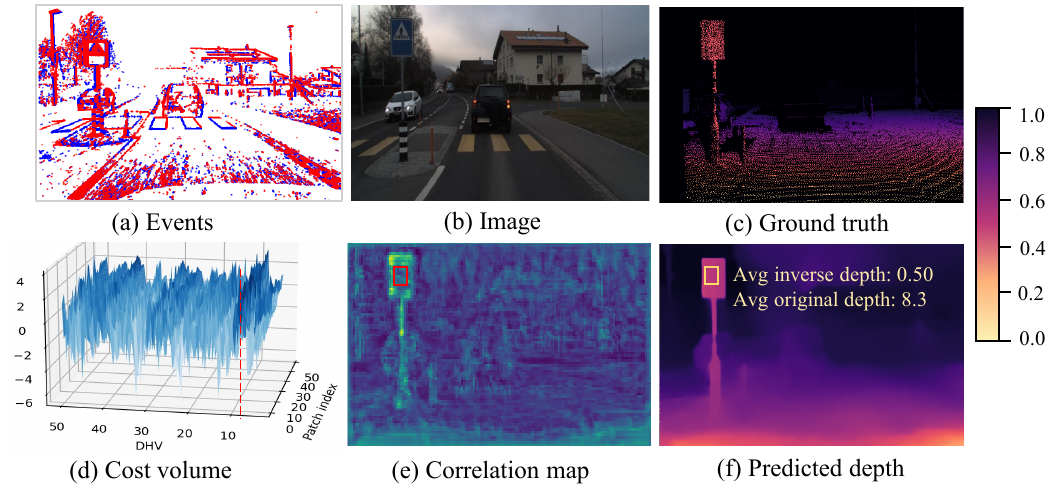}
\caption{Visualization of the correspondence between contextual features and DHV embedding features. (d) shows the cost volume distribution of the patch in (f), where the depth value of the DHV  at the highest similarity point closely matches the predicted depth, and neighboring pixels exhibit similar distributions. (e) visualizes the correlation map, revealing clear spatial structures.}
\label{fig_1}
\end{figure}

To fully leverage the complementary strengths of events and images, significant research efforts~\cite{unict, RAMNet, pcdepth, srfnet} have focused on cross-modal alignment and fusion techniques. Typically, asynchronous event streams are transformed into grid-like representations for feature-level integration with conventional images. Specifically, RAMNet~\cite{RAMNet} employs RNNs to align events and images, while UniCT~\cite{unict} uses self-attention, and works~\cite{pcdepth, srfnet} iteratively refine feature fusion patterns to alleviate representation bottlenecks. Despite these advances, existing methods remain limited by their primary focus on improving contextual features. They fundamentally struggle with the inherent ill-posedness of MDE, where directly regressing the full depth distribution remains highly nonlinear and challenging.

To overcome this limitation, we introduce a novel iterative refinement paradigm for event-image MDE that fundamentally reformulates the ill-posed depth regression as a constrained depth search task. The core of our approach is a \textbf{Depth Hypothesis Volume (DHV)}, which discretizes the continuous depth space into a set of candidate depth values within adaptive bounds derived from prior estimates. This discretization provides inherent robustness by constraining the solution space to a structured set of plausible depth hypotheses, effectively mitigating the impact of prediction noise and enabling stable residual optimization. Conceptually aligned with search-based optical flow methods such as RAFT~\cite{raft}, which maintains a 4D cost volume built from two frames and iteratively predicts residual flow by narrowing the 2D search space, we construct a 3D cost volume to encode matching confidence between the DHV embeddings and context features, as shown in Fig.~\ref{fig_1}(d, f). Operating across multiple scales, the cost volume captures both distant layout and fine geometric details, providing geometrically meaningful guidance for iterative refinement. 

\begin{figure*}[t]
\centering
\includegraphics[width=0.95\textwidth,height=0.25\textwidth]{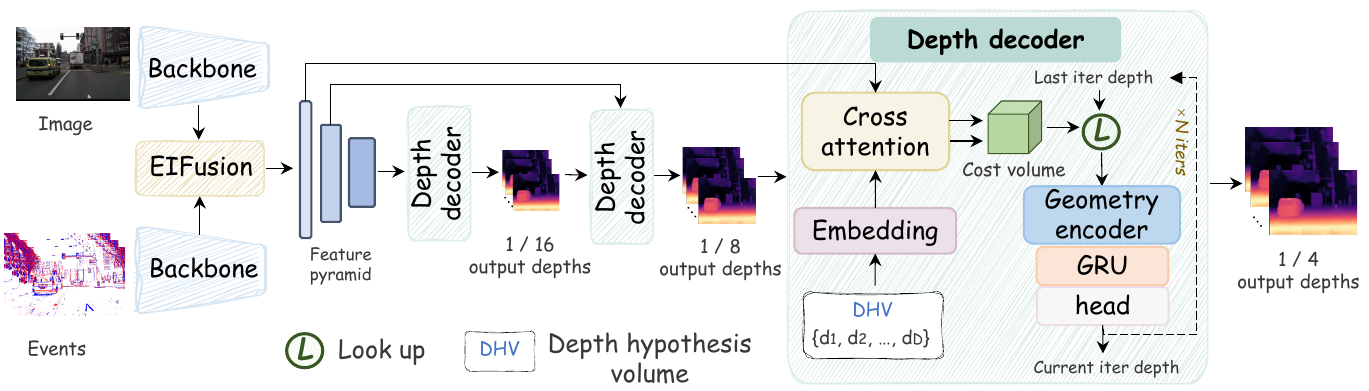}
\caption{Overview of our proposed \textbf{HypoDepth}. The network architecture comprises three main components: 1) \textbf{Backbone}: The image and event backbones respectively extract multi-resolution contextual features from image frames and event streams. 2) \textbf{EIFusion}. The fusion module adaptively fuses information from both modalities across multiple resolutions to produce unified contextual representations. 3) \textbf{Depth Decoder}. This decoder constructs a Depth Hypothesis Volume (DHV) and computes a 3D cost volume between DHV and the fused contextual features for lookup and geometry encoding. Subsequently, a GRU-based iterative unit progressively predicts residual depth and finally regresses the complete depth map.}
\label{fig_2}
\end{figure*}

Building on this design, we develop \textbf{HypoDepth} (\cref{fig_2}), a framework that establishes an iterative refinement paradigm for event-image monocular depth estimation. To effectively balance computational efficiency with multi-scale representation, we adopt resolution-specific cross-modal fusion strategies for contextual feature extraction. For each resolution, a globally shared DHV is first established using prior depth estimates and projected into a unified feature space to represent the scene geometry under different depth hypotheses. We then construct the 3D cost volume by computing similarity metrics between the contextual features and the DHV embeddings, followed by multi-scale correlation lookups to integrate structures across varying distances. Since the DHV provides discrete depth, a geometry encoder is introduced to incorporate previous depth estimates and contextual features to produce a spatially consistent geometric representation (\cref{fig_1}~(e)). This feature is then processed by a GRU-based iterative unit to predict residual depths, achieving progressive refinement through multiple iterations. Benefiting from the lightweight 3D cost volume, this process can be efficiently operated across resolutions, enabling effective global-to-local depth refinement across various resolutions. 
Extensive experiments on DSEC~\cite{dsec} and MVSEC~\cite{mvsec} demonstrate our state-of-the-art performance. Our main contributions are as follows:
\begin{itemize}
         \item We propose \textbf{HypoDepth}, a novel iterative refinement framework for event-image monocular depth estimation. By introducing a DHV, it reformulates ill-posed depth regression into a constrained depth search task, enabling stable residual depth optimization.
         \item We construct efficient multi-resolution 3D cost volumes coupled with multi-scale correlation lookup, which allows comprehensive geometric feature extraction for depth refinement from global structure to local details.
         \item Our approach achieves the best performance on DSEC and MVSEC datasets with superior generalization ability. 
\end{itemize}

\section{Related Work}
\label{sec: Related Work}

\textbf{Image-based Monocular Depth Estimation.} 
Inspired by the success of deep learning, early CNN-based methods~\cite{eigen2014depth, laina2016deeper} in the image domain demonstrate that neural networks can infer monocular scene geometry. DORN~\cite{fu2018deep} and BTS~\cite{BTS} further improve accuracy through depth discretization and scale-aware regression. LapDepth~\cite{mde_residuals} introduces a Laplacian residual pyramid to guide depth completion, achieving coarse-to-fine depth refinement. Leveraging strong self-attention, transformer-based models~\cite{bhat2021adabins, ranftl2020towards, DPT, yang2024depth, yang2024depth2} achieve remarkable progress. AdaBins~\cite{bhat2021adabins} introduces adaptive and learnable depth bins for fine-grained and scale-consistent estimation, while works~\cite{ranftl2020towards, DPT, yang2024depth, yang2024depth2} enhance generalization through large-scale pretraining and self-distillation. RayFusion~\cite{rayfusion} fuses ray-based cost volumes across multiple frames to achieve consistent and accurate sparse depth completion. Recently, diffusion-based methods~\cite{shao2024monodiffusion, duan2024diffusiondepth} reformulate depth prediction as generative tasks, producing more detailed depth maps. However, these approaches often fail under motion blur or poor illumination, where event cameras can still provide reliable information.

\noindent\textbf{Event-based Monocular Depth Estimation.} 
Inspired by image-based learning methods, some works convert events into image-like representations for data-driven prediction. Early event-based method~\cite{e2depth} first introduces recurrent units into a U-Net to encode temporal consistency, while HMNet~\cite{HMNet} and EReFormer~\cite{EReformer} capture spatiotemporal relations using multi-scale strategies and recursive Transformers, respectively. Self-supervised approaches~\cite{tespec, zhu2023self} aim to reduce depth annotation and model long-term information. DepthAnyEvent~\cite{depthanyevent} further distills geometric priors from image models to improve generalization. To compensate for missing appearance, RAMNet~\cite{RAMNet} first fuses events' temporal cues and image semantics for cross-modal depth estimation. SRFNet~\cite{srfnet} and PCDepth~\cite{pcdepth} perform iterative feature fusion refinement, whereas UniCT~\cite{unict} and FUSE~\cite{fuse} employ transformer or hybrid attention for enhanced alignment. Although these methods strengthen contextual representations for MDE, their direct depth regression remains highly nonlinear and difficult to converge due to the ill-posed nature of monocular estimation.

\noindent\textbf{Iterative Refinement Scheme.} 
In event-based MDE, PCDepth~\cite{pcdepth} adopts iterative strategies to optimize feature-level fusion rather than explicit depth refinement. RAFT~\cite{raft} introduces a cost-volume-based iterative refinement scheme to optical flow estimation that greatly improves accuracy by predicting residual flows. This paradigm is later extended to stereo matching~\cite{raftstereo, xu2025igev++} and event-based flow estimation~\cite{eraft, TMA, edcflow}, achieving state-of-the-art performance. Unlike these approaches that rely on explicit viewpoint correspondences, we introduce a discrete Depth Hypothesis Volume (DHV) to simulate such correspondences between the DHV features and contextual features, enabling iterative local search and explicit geometric refinement through cost-volume construction.

\section{Method}
\label{sec:method}
Given an image $I$ and its corresponding event stream $\mathcal{E}$, we aim to estimate a dense depth map. Inspired by RAFT's~\cite {raft} cost-volume-based iterative refinement, we propose \textbf{HypoDepth}, which constructs a discrete Depth Hypothesis Volume (DHV) in a virtual depth space to simulate matching and build a 3D cost volume for iterative refinement, providing explicit constraints to alleviate the ill-posed nature of MDE. As shown in~\cref{fig_2}, HypoDepth comprises image–event encoders, multimodal fusion modules, and iterative depth decoders with multi-resolution to regress depth from coarse to details. The following sections detail the event representation and key components of HypoDepth.

\subsection{Event Representation}
Event cameras asynchronously generate events by detecting brightness changes at each pixel. Each event is represented as $(x, y, t, p)$, where $(x, y)$, $t$, and $p\in\{-1, +1\}$ denote the spatial coordinate, timestamp, and polarity of the brightness change, respectively. To make the event data compatible with neural network inputs, we convert the event stream $\mathcal{E}$ into an image-like voxel grid~\cite{zhu_unflow} with $B$ temporal bins:
\begin{align}
\mathcal{V}\left(b, x, y\right)&=\sum_{j=0}^{N_e-1} p_{j}\max \left(0,1-\left|b-t_{j}^{*}\right|\right),\label{eq1}\\
t_{j}^{*} &= \frac {(t_{j} - t_{min})}{t_{max} - t_{min}} \times (B - 1),\label{eq2}
\end{align}
where $b\in[0, B-1]$ is the bin index, $N_e$ is the number of events, and $t_{max}$ and $t_{min}$ are the maximum and minimum timestamps, respectively.

\subsection{Cross-Modal Context Encoder}
\label{sec:FeatureEX}
\noindent\textbf{Backbone.}
We employ fine-tuned Swin-T~\cite{liu2021swin} as our encoder, leveraging its strong representations pre-trained on large-scale datasets. To adapt the event voxel grid with its $B$ temporal bins to the encoder, we prepend several convolutional layers to project it into a 3-channel input. Given the modality gap between images and events, we use separate encoders without weight sharing, from which we extract multi-resolution features at $1/16$, $1/8$, and $1/4$ resolutions, denoted as $F_I^{j}$ and $F_E^{j}$ for $j \in \{1/16, 1/8, 1/4\}$.

\begin{figure}[t]
  \centering
  \includegraphics[width=0.9\linewidth, height=0.57\linewidth]{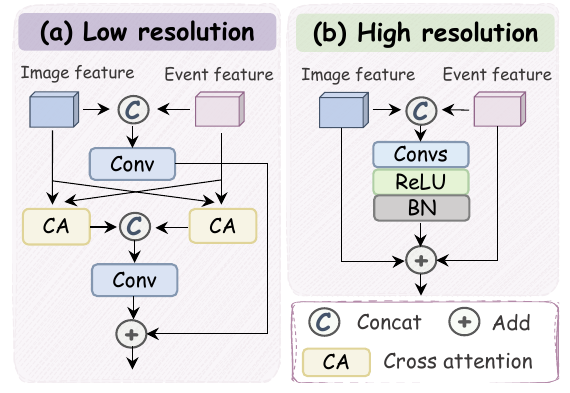}
\caption{Illustration of EIFusion module. (a) The cross-attention-based fusion strategy for low-resolution (1/16 and 1/8 resolution) features. (b) The efficient convolution-based fusion method for high-resolution (1/4 resolution) features.}
\label{eifusion}
\end{figure}

\noindent\textbf{EIFusion.}
Images provide rich appearance information but tend to fail under challenging conditions, while events offer valuable texture cues in such environments but lack rich appearance features. To fully exploit their complementarity, we perform adaptive multimodal fusion rather than simple concatenation. Considering computational efficiency and the receptive field differences across resolutions, we apply different fusion strategies. At the 1/16 and 1/8 levels, cross-attention (CA) is employed to achieve global semantic alignment. As illustrated in~\cref{eifusion}~(a), we first concatenate the two feature maps and apply a convolution for implicit alignment and channel restoration, which serves as a skip connection. Then, $F_I$ and $F_E$ are respectively used as $Q$, $K$, and $V$ inputs to a shared CA module to enable global information interaction:
\begin{align}
F^{\prime}_I &= F_I + \text{CA}(Q_I, K_E, V_E), \label{eq3}\\
F^{\prime}_E &= F_E + \text{CA}(Q_E, K_I, V_I), \label{eq4}
\end{align}
where $Q$, $K$, and $V$ are obtained by linearly projecting $F_I$ and $F_E$ through MLPs. The aligned features are then concatenated to form a fused representation, followed by a convolution layer to restore the channel dimension, which is added to the skip connection to obtain $F_{ct}^{j}$, where $j=\{1/16, 1/8\}$. For the 1/4 features, which focus on structural details, convolution-based fusion is adopted. As shown in~\cref{eifusion}~(b), the features from both modalities are concatenated and fused locally through convolution, followed by ReLU and batch normalization: 
\begin{align}
F^{\prime} &= \text{BN}(\text{ReLU}(\text{Convs}(\text{Concat}(F_I, F_E))), \label{eq5}
\end{align}
Skip connections are then added to prevent information loss, yielding $F_{ct}^{1/4}$. The multi-resolution fused contextual features $F_{ct}^{j}, j=\{1/16, 1/8, 1/4\}$ are then used for subsequent cost volume construction and depth regression.

\subsection{Iterative Depth Decoder}
\label{sec:decoder}
To achieve depth residual optimization, we develop a 3D cost volume construction pipeline based on a depth hypothesis volume (DHV) and progressively narrow the depth search range for multi-step refinement. In the following, we take a resolution as an example to illustrate the iterative decoder and denote the contextual features as $F_{ct} \in \mathbb{R}^{c \times H \times W}$, where $c$, $H$, and $W$ denote the number of channels, height, and width, respectively.

\noindent\textbf{Depth Hypothesis Volume (DHV).} 
To simulate the matching process, we introduce a global DHV. Specifically, we uniformly sample $D$ depth values within the prior minimum depth $d_{min}$ and maximum depth $d_{max}$ range to form the depth candidates for each pixel, \textit{i.e.}, $\{\hat{d}_i\}_{i=0}^{D-1}$. Similar to RAMNet~\cite{RAMNet}, for improved robustness, our network predicts normalized log depth. Therefore, we convert the $D$ sampled depth values into normalized log depths to construct the DHV:
\begin{align}
DHV &= \{d_0, d_1, \cdots, d_{D-2}, d_{D-1}\}. \label{eq6} \\
d_i &= \frac{\text{log}(\hat{d}_i)-\text{log}(d_{min})}{\text{log}(d_{max})-\text{log}(d_{min})}, \label{eq7}
\end{align}
To establish correspondence between the contextual features and the DHV, we embed the DHV into the same feature space as $F_{ct}$ by linear projection:
\begin{align}
F_{dhv} = \text{Embedding}(DHV), \label{eq8}
\end{align}
where $F_{dhv} \in \mathbb{R}^{c \times D}$ reflects the geometric projection of the scene under each specific depth hypothesis.

\begin{figure}[t]
  \centering
  \includegraphics[width=0.85\linewidth, height=0.76\linewidth]{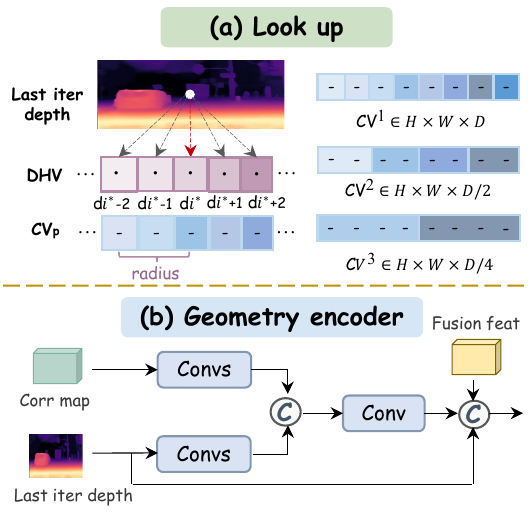}
\caption{Illustration of correlation lookup and geometry encoder. (a) Cost volume lookup with correlation pooling. The red arrow indicates the index of the depth value in the DHV that is closest to the previous iteration’s depth updation at pixel $P$. This position serves as the center for the current correlation lookup $C_P$ of the pixel $P$. (b) Geometry feature encoder for depth update.}
\label{lookup}
\end{figure}

\noindent\textbf{3D Cost Volume.} As in~\cref{fig_2}, to provide more stable geometric associations for the cost volume, we first perform bidirectional cross-attention (CA) between the contextual features $F_{ct}$ and the DHV-embedded features $F_{dhv}$:
\begin{align}
F^{\prime}_{dhv} &= F_{dhv} + \text{CA}(Q_{dhv}, K_{ct}, V_{ct}), \label{eq9} \\
F^{\prime}_{ct} &= F_{ct} + \text{CA}(Q_{ct}, K_{dhv}, V_{dhv}). \label{eq10}
\end{align}
Using $F_{dhv}$ and $F_{ct}$ as each other’s queries ($Q$), the contextual features gain perception of geometric information under different depth hypotheses, while DHV embeddings are enhanced with scene semantic cues. We then compute the similarity between these two features to construct the 3D cost volume:
\begin{align}
CV = \text{Sim}(F^\prime_{ct}, F^{\prime}_{dhv}) \in \mathbb{R}^{H \times W \times D}, \label{eq11}
\end{align}
where $\text{Sim}(\cdot)$ denotes the dot-product. The $CV$ encodes the matching confidence across different depth hypotheses along the depth dimension, serving as the basis for depth refinement. We further pool $CV$ along the depth dimension to construct a cost volume pyramid $\{CV^1, CV^2, CV^3\}$, providing multi-scale depth cues for both nearby and distant objects, as illustrated in~\cref{lookup} (a).

\noindent\textbf{Correlation Lookup.} 
At the $k$-th iteration, we perform a local search along the depth dimension of $CV$ to retrieve current correlation information. As illustrated in~\cref{lookup} (a), given the $(k-1)$-th estimated depth map $\Psi^{k-1}$, for each pixel $P$, we compute the Manhattan distance between its depth value $\psi_P^{k-1}$ and all depth hypotheses $\{d_i\}_{i=0}^{D-1}$ in the DHV. The position with the smallest distance is selected as the lookup center for pixel $P$ in the current iteration:
\begin{equation}\label{eq12}
i^\ast = \arg\min\limits_{i}|\psi_P^{k-1} - d_i|, {i \in \{0,1,\dots,D-1\}}.
\end{equation}
In the cost volume $CV_P$ of pixel $P$, we perform a local search with a radius $r$ to obtain the correlation map:
\begin{equation}
Corr(P) = \big\{\, CV_P(i^\ast + \Delta i) \;\big|\; \Delta i \in [-r, r] \,\big\} \in \mathbb{R}^{2r+1}. \label{eq13}
\end{equation}
For the multi-scale cost volumes, the lookup centers are determined from the corresponding pooled $\{DHV^l\}_{l=1}^3$, and the resulting correlation maps are concatenated to form a multi-scale representation $Corr \in \mathbb{R}^{3\ast(2r+1) \times H \times W}$.

\noindent\textbf{Depth Update.} Before regressing depth, the correlation features are processed by a geometry encoder. As shown in~\cref{lookup} (b), the multi-scale correlation $Corr$ is processed by two convolutional layers and then fused with the depth features via concatenation. We further inject the fused multi-modal contextual features $F_{ct}$ to provide structural scene information. The generated final geometric feature is fed into an update unit with a GRU-based iterative mechanism and a prediction head to estimate the residual depth $\Delta \Psi$. The updated depth at $k$-th iteration is then computed as: $\Psi^{k} =\Psi^{k-1} + \Delta \Psi$, which is then upsampled to the original resolution using convex upsampling~\cite{raft} and scaled to real-world depth values for loss computation:
\begin{equation}
\hat{\Psi}^k = d_{max}\text{exp}(\text{log}\frac{d_{max}}{d_{min}}(\Psi ^k -1)). \label{eq14}
\end{equation}

\subsection{Multi-resolution Depth Refinement}
\label{subsec:multi-resolution}
In our 3D correlation, the depth dimension $D \ll H \times W$ allows efficient refinement at higher resolutions. We adopt a coarse-to-fine process from $1/16$ to $1/4$ resolution: low-resolution stages emphasize global consistency, while high-resolution stages focus on recovering fine structural details. To adapt to different refinement granularity, the number of depth hypotheses $D$ in the DHV is adjusted accordingly, a smaller one for lower resolutions and a larger one for higher ones. The initial depth at $1/16$ stage is set to zero.

\subsection{Loss}
\label{subsec:loss}
Following RAFT~\cite{raft} and PCDepth~\cite{pcdepth}, we supervised our network with $SI_{log}$ loss~\cite{eigen2014depth} between ground truth $\Psi^{gt}$ and predictions $\{\hat{\Psi}^i\}_{i=1}^K$, with increasing weights:
\begin{align}
\mathcal{L} &= \sum_{i=1}^{K} 0.8^{K-i} \alpha \sqrt{\mathbb{V}(\delta _i) - \lambda \mathbb{E}(\delta _i)}, \label{eq15} \\ 
\delta _i &= \text{log}(\hat{\Psi}^i) - \text{log}(\Psi^{gt}), \label{eq16}
\end{align}
where $\mathbb{V}(\cdot)$ and $\mathbb{E}(\cdot)$ denote variance and expectation operations, respectively, and $\alpha$ and $\lambda$ are set to 10 and 0.15.
\section{Experiments}
\label{sec:Experiments}
\begin{table*}[t]
    \centering
    \caption{Evaluation on DSEC~\cite{dsec}. The best results are in bold, while the second-best ones are underlined. $^{\dagger}$ and $^{\ddagger}$ denote the result from DepthAnyEvent~\cite{depthanyevent}and this paper, respectively. `E' and `E+I' denote event-only and event–image inputs, respectively.} 
    \label{dsec} 
    \resizebox{0.82\linewidth}{!}{
    \begin{tabular}{ccccccccccc}
\toprule 
        Input & Method & $\delta$1 $\uparrow$ & $\delta$2 $\uparrow$ & $\delta$3 $\uparrow$ & Abs Rel $\downarrow$ & RMSE $\downarrow$ & RMSE log $\downarrow$ & FLOPs (G) & Param (M) & Infer time (ms) \\ 
\midrule
        \multirow{8}{*}{E} & E2Depth$^{\dagger}$~\cite{e2depth} & 0.447 & 0.755 & 0.915 & 0.344 & 13.467 & 0.376 & 93 & 10.7 & 2.3 \\
        ~ & EReformer$^{\dagger}$~\cite{EReformer} & 0.486 & 0.776 & 0.892 & 0.387 & 13.954 & 0.395 & 36 & 41.8 & 50.1 \\
        ~ & EReformer-C$^{\ddagger}$~\cite{EReformer} & 0.463 & 0.755 & 0.899 & 0.399& 8.635 & 0.382 & 109 & 41.8 & 144.3 \\
        ~ & DepthAnyEvent~\cite{depthanyevent} & 0.585 & 0.860 & 0.955 & 0.277 & 11.117 & 0.292 & 33 & 24.2 & 1.8 \\
        ~ & DepthAnyEvent-R~\cite{depthanyevent} & 0.592 & 0.900 & 0.971 & 0.252 & 9.824 & 0.268 & 39 & 25.5 & 12.9 \\
        ~ & \textbf{Ours-E-T} & 0.755 & 0.934 & 0.995 & 0.168 & 5.377 & 0.215 & 5 & 1.4 & 11.8 \\
        ~ & \textbf{Ours-E-S} & 0.828 & 0.960 & 0.988 & 0.137 & 4.505 & 0.176 & 34 & 29.6 & 33.0 \\
        ~ & \textbf{Ours-E-B} & 0.839 & 0.962 & 0.991 & 0.129 & 4.442 & 0.170 & 134 & 32.3 & 58.2 \\
\midrule   
        \multirow{3}{*}{E + I} & PCDepth$^{\ddagger}$~\cite{pcdepth} & \underline{0.893} & 0.980 & \underline{0.996} & 0.103 & 3.591 & 0.137 & 162 & 66.5 & 64.3 \\ 
        ~ & \textbf{Ours-B} & \textbf{0.901} & \underline{0.982} & \textbf{0.997} & \underline{0.099} & \underline{3.583} & \underline{0.135} & 155 & 60.2 & 71.6 \\
        ~ & \textbf{PCDepth-Ours} & \textbf{0.901} & \textbf{0.983} & \textbf{0.997} & \textbf{0.098} & \textbf{3.559} & \textbf{0.134} & 209 & 68.6 & 73.7 \\
\bottomrule 
    \end{tabular}}
\end{table*}

\begin{figure*}[h]
\centering
\includegraphics[width=0.84\textwidth,height=0.28\textwidth]{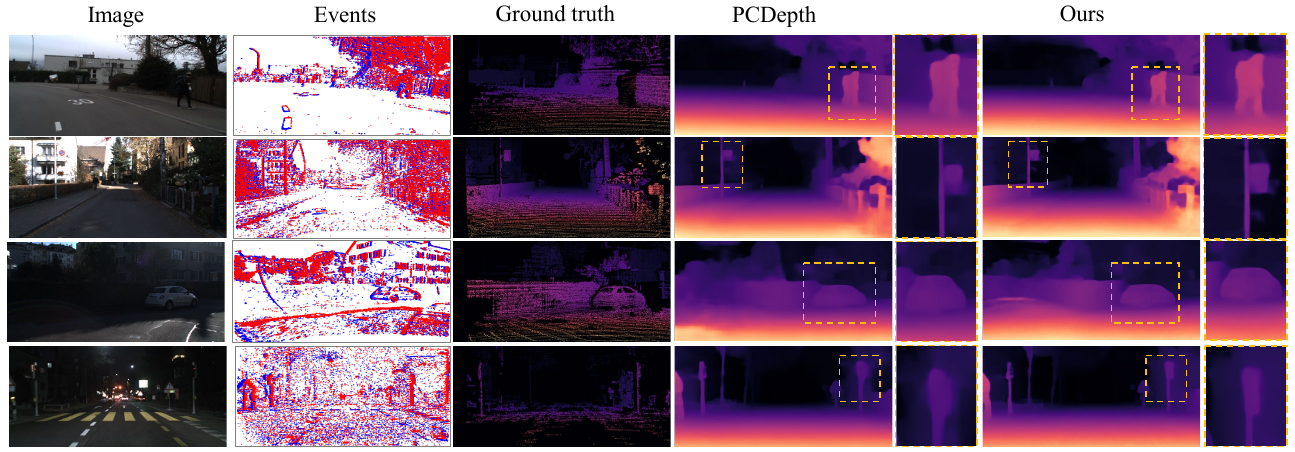}
\caption{\textbf{Qualitative results of monocular depth predictions on DSEC~\cite{dsec}.} Notable areas are marked by bounding boxes.}
\label{dsec_viz}
\end{figure*}

\noindent\textbf{Datasets and metrics.} We conduct experiments on two real-world datasets, MVSEC~\cite{mvsec} and DSEC~\cite{dsec}. MVSEC provides outdoor sequences under day and night conditions with a resolution of $260\times346$ and a depth range from 1.97 to 80.0 meters, featuring relatively simple scenes and sparse events. Following previous works~\cite{HMNet, RAMNet, pcdepth}, we pretrain our model on the synthetic dataset EventScape~\cite{RAMNet}, fine-tune on $outdoor\_day2$, and evaluate on $outdoor\_day1$ and $outdoor\_night1$. DSEC is a more complex and diverse outdoor dataset with higher resolution ($480 \times 640$), denser events, and a depth range of 4.53 to 50.1 meters. We convert the provided disparity ground truth to depth maps using the official code. Since ground truth for the test set is not publicly available, we follow works~\cite{bartolomei2024lidar, depthanyevent} to split the training set into new train/test subsets, and apply a $320 \times 640$ center crop to reduce misalignment artifacts. We utilize absolute relative error (Abs Rel), root mean squared error (RMSE), logarithmic RMSE (RMSE log), and accuracy $\delta n$ ($\delta<1.25^n$, $n = 1, 2, 3$) to evaluate the predicted depth. For DSEC, we also report Floating-Point Operations (FLOPs, G), model size (Param), and inference time (Infer time) under the same computational environment.

\noindent\textbf{Implementation Details.} For MVSEC and DSEC, we use event slices of 50 ms and 20 ms, respectively, with the number of temporal bins $B=3$. The iteration numbers per stage are set to 8. The lookup radius $r$ is 8, and the depth hypothesis numbers $D$ for different resolutions are (32, 64, 96). Detailed network configurations are provided in the supplementary material. Our framework is implemented in PyTorch, and all experiments are trained with the AdamW optimizer~\cite{adamw} and the one-cycle learning rate scheduler~\cite{onecycle}. For DSEC, we train for 60k iterations with a maximum learning rate of 2e-4 and a batch size of 16. For MVSEC, we fine-tune from a model pretrained on EventScape for 40k iterations using a maximum learning rate of 1e-5 with a batch size set to 24.

\subsection{Results}
\noindent\textbf{DSEC Dataset.}
We report the results on DSEC~\cite{dsec} in~\cref{dsec}, including our fusion-based and event-only versions. To provide comparisons between accuracy and efficiency, we evaluate three model variants: base (B), small (S), and tiny (T). They differ in feature dimension, number of iterations, and the $D$ in the DHV. We reimplement PCDepth~\cite{pcdepth} from scratch using the official code, while \textbf{PCDepth-Ours} cascades our iterative refinement of PCDepth. For a fair comparison, we retrain EReFormer~\cite{EReformer} by applying its recurrent strategy only to the current event stream, following the original paper’s setting, and denote this model as EReFormer-C.

All versions of our method consistently outperform event-only methods~\cite{e2depth, EReformer, depthanyevent}, with a 49\% improvement in Abs Rel (\textbf{Ours-E-B} vs. DepthAnyEvent-R~\cite{depthanyevent}). This gain mainly contributes to two factors: (1) these U-Net-like methods downsample inputs to very low resolutions and recover details through upsampling and skip connections, which inevitably causes information loss. In contrast, our approach regresses depth from low to high resolutions, maintaining both global consistency and fine structural details; (2) the proposed Depth Hypothesis Volume (DHV) constrains the search space for the optimal solution, while the cost-volume-based iterative refinement gradually converges to it, leading to substantial accuracy improvement. Moreover, the fusion of event and image provides a significant benefit on this more complex DSEC dataset~\cite{dsec}. Compared with fusion-based approaches, our method surpasses PCDepth~\cite{pcdepth}, which focuses on feature-level fusion rather than direct depth refinement. When integrating our refinement strategy with PCDepth (\textbf{PCDepth-Ours}), further performance gains are observed across all metrics. 

Since our method is fusion-based, we qualitatively compare it with PCDepth~\cite{pcdepth} in~\cref{dsec_viz}. Our approach shows superior robustness in dynamic scenes (e.g., the pedestrian in row 1), better structural detail (rows 2–4), and finer depth refinement across near and far regions. This superiority mainly arises from our multi-scale cost-volume lookup, which balances depth estimation across different ranges, and coarse-to-fine refinement, which enhances depth precision, whereas PCDepth performs at a single resolution.

\begin{table*}[t]
    \centering
    \caption{Evaluation on MVSEC~\cite{mvsec}. The best results are in bold, while the second-best ones are underlined.} 
\label{mvsec} 
    \resizebox{0.9\linewidth}{!}{
    \begin{tabular}{cccccccccccccc}
\toprule 
        \multirow{2}{*}{Input} & \multirow{2}{*}{Method} & \multicolumn{6}{c}{Outdoor\_day1} & \multicolumn{6}{c}{Outdoor\_night1} \\ 
        \cmidrule{3-14}
        ~ & ~ & $\delta$1 $\uparrow$ & $\delta$2 $\uparrow$ & $\delta$a3 $\uparrow$ & Abs Rel $\downarrow$ & RMSE $\downarrow$ & RMSE log $\downarrow$ & $\delta$1 $\uparrow$ & $\delta$2 $\uparrow$ & $\delta$3 $\uparrow$ & Abs Rel $\downarrow$ & RMSE $\downarrow$ & RMSE log $\downarrow$  \\ 
\midrule
        \multirow{3}{*}{E} & E2Depth~\cite{e2depth} & 0.567 & 0.772 & 0.876 & 0.346 & 8.564 & 0.421 & 0.408 & 0.615 & 0.754 & 0.591 & 11.21 & 0.646 \\
        ~ & EReformer~\cite{EReformer} & 0.664 & 0.831 & 0.923 & 0.271 & - & 0.333 & 0.547 & 0.753 & 0.881 & 0.317 & -  & 0.415  \\
        ~ & HMNet~\cite{HMNet} & 0.690 & 0.849 & 0.930 & 0.254 & 6.890 & 0.319 & 0.513 & 0.714 & 0.837 & 0.323 & 9.008 & 0.482  \\
        
\midrule  
        \multirow{5}{*}{E + I} &  SRFNet~\cite{srfnet} & 0.637 & 0.810 & 0.900 & 0.268 & 8.453 & 0.375 & 0.433 & 0.662 & 0.8 & 0.371 & 11.469 & 0.521 \\
        ~ & RAMNet~\cite{RAMNet} &  0.541 & 0.778 & 0.877 & 0.303 & 8.526 & 0.424 & 0.296 & 0.502 & 0.635 & 0.583 & 13.340 & 0.830 \\
         ~ & HMNet~\cite{HMNet} & \textbf{0.717} & 0.868 & 0.940 & 0.230 & 6.922 & 0.310 & 0.497 & 0.661 & 0.784 & 0.349 & 10.818 & 0.543 \\
        ~ & PCDepth~\cite{pcdepth} & \underline{0.712} & 0.867 & 0.941 & \underline{0.228} & \textbf{6.526} & \underline{0.301} & \textbf{0.632} & \textbf{0.822} & \underline{0.922} & \underline{0.271} & \textbf{6.715} & \textbf{0.354} \\ 
        ~ & \textbf{Ours-B} & 0.698 & \textbf{0.874} & \textbf{0.948} & \textbf{0.212} & \underline{6.556} & \textbf{0.296} & \underline{0.614} & \underline{0.818} & \textbf{0.923} & \textbf{0.268} & \underline{6.834} & \underline{0.355} \\ 
\bottomrule 
    \end{tabular}}
\end{table*}

\noindent\textbf{MVSEC Dataset.} 
~\cref{mvsec} compares the performance of existing methods on the MVSEC~\cite{mvsec}. \textbf{Ours-B} achieves SOTA results on both day and night sequences in terms of the main metric Abs Rel. Event-only methods EReFormer~\cite{EReformer} and HMNet~\cite{HMNet} benefit from modeling long-term dependencies and outperform the fusion-based RAMNet~\cite{RAMNet}, but they remain limited by the U-Net architecture, the lack of appearance cues, and the absence of depth-level refinement. SRFNet~\cite{srfnet} and PCDepth~\cite{pcdepth} enhance details at higher resolutions through optimizing multimodal fusion, yet their refinement remains static at the feature level without explicit constraint or optimization in the depth space. In contrast, our method performs dynamic, depth-level iterative refinement via a cost-volume-based lookup mechanism, progressively refining depth from low to high resolutions while mitigating the ill-posedness of monocular estimation.


\begin{table}[htbp]
    \centering
    \caption{Zero-shot Generalization on MVSEC~\cite{mvsec} and DSEC~\cite{dsec}. All methods are trained on EventScape~\cite{RAMNet}. The best are in bold.} 
\label{zero-shot} 
    \resizebox{1.0\linewidth}{!}{
    \begin{tabular}{cccccccc}
\toprule
        Dataset & Method & $\delta$1 $\uparrow$ & $\delta$2 $\uparrow$ & $\delta$3 $\uparrow$ & Abs Rel $\downarrow$ & RMSE $\downarrow$ & RMSE log $\downarrow$ \\ 
\midrule 
        \multirow{3}{*}{\thead{MVSEC \\ outdoor\_day1}} & EReFormer-C~\cite{EReformer}  & 0.157 & 0.256 & 0.348 & 0.580 & 14.510 & 1.336 \\
        ~ & PCDepth~\cite{pcdepth}  & 0.226 & 0.446 & 0.607 & 0.471 & 12.650 & 0.874 \\
        ~& \textbf{Ours-B}  & \textbf{0.248} & \textbf{0.486} & \textbf{0.640} & \textbf{0.432} & \textbf{12.480} & \textbf{0.809} \\
\midrule   
        \multirow{3}{*}{\thead{MVSEC \\ outdoor\_night1}} & EReFormer-C~\cite{EReformer}  & 0.117 & 0.216 & 0.300 & 0.621 & 16.35 & 1.472 \\
        ~ & PCDepth~\cite{pcdepth} & 0.197 & 0.400 & 0.591 & 0.544 & 12.820 & 0.795 \\
        ~& \textbf{Ours-B} & \textbf{0.241} & \textbf{0.471} & \textbf{0.653} & \textbf{0.490} & \textbf{12.300} & \textbf{0.723} \\
\midrule   
        \multirow{3}{*}{DSEC} & EReFormer-C~\cite{EReformer}  & 0.018 & 0.092 & 0.182 & 0.840 & 21.970 & 2.216 \\
         ~ & PCDepth~\cite{pcdepth} & 0.142 & 0.290 & 0.450 & 0.511 & 16.660 & 0.889 \\
        ~& \textbf{Ours-B} & \textbf{0.145} & \textbf{0.294} & \textbf{0.454} & \textbf{0.487} & \textbf{15.95} & \textbf{0.855} \\
\bottomrule 
    \end{tabular}}
\end{table}

\noindent\textbf{Zero-shot Generalization.} 
We pre-train our model on the synthetic dataset EventScape~\cite{RAMNet} and perform zero-shot transfer to DSEC~\cite{dsec} and MVSEC~\cite{mvsec} to assess generalization, as shown in~\cref{zero-shot}. Under the same settings, we reproduce EReFormer-C~\cite{EReformer} and PCDepth~\cite{pcdepth} for comparison. Our method achieves the best performance across all scenes, particularly on the complex DSEC. The superior generalization and sim-to-real transfer come from the scene-adaptive depth hypothesis and the multi-resolution refinement enabled by multi-scale cost-volume lookups.

\begin{figure}[t]
\centering
\includegraphics[width=0.4\textwidth,height=0.24\textwidth]{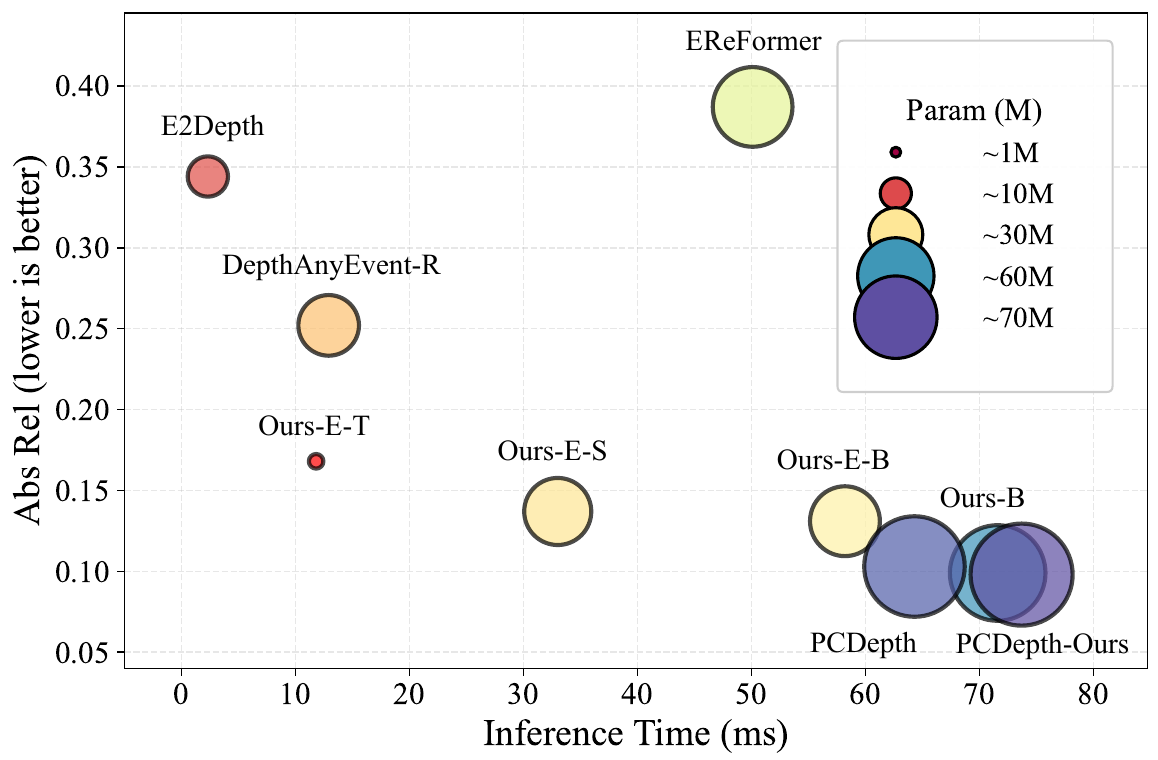}
\caption{Abs Rel vs. Inference time (ms) and model size (Param, M). All models are tested with one NVIDIA 4090 GPU. The size of the circle corresponds to the model size, as shown in the legend.}
\label{performance}
\end{figure}

\noindent\textbf{Model Efficiency.}
In~\cref{dsec}, we also compare the model efficiency. Both our method and PCDepth~\cite{pcdepth} perform iterative refinement at high resolutions, leading to relatively high computations and inference time. However, benefiting from our lightweight 3D cost volume, the computation across multiple resolutions remains lower than that of PCDepth. Moreover, \textbf{Ours-E-T} achieves the best balance between efficiency metrics and accuracy, as illustrated in~\cref{performance}. Although E2Depth~\cite{e2depth} and DepthAnyEvent~\cite{depthanyevent} run faster, they sacrifice accuracy. Therefore, our Tiny model is well-suited for real-time applications or computationally constrained scenarios.

\subsection{Ablation Study}
We conduct ablation studies on a scaled-down model of our base model using DSEC~\cite{dsec}, as shown in~\cref{ablation}.

\noindent\textbf{EIFusion.} 
We replace the EIFusion with simple concatenation, which prevents the effective exploitation of the complementarity of images and events, leading to insufficient contextual features and blurred cost volume matching.

\noindent\textbf{W/o CA in Decoder.} 
We remove the cross-attention between the DHV embeddings and the fused contextual features to verify its necessity. The results show that this weakens their semantic and geometric correspondence and reduces the accuracy of the cost volume.

\begin{table*}[t]
    \centering
    \caption{Ablation studies on DSEC~\cite{dsec} with scaled-down model. } 
\label{ablation} 
    \resizebox{0.8\linewidth}{!}{
    \begin{tabular}{cccccccccc}
\toprule 
        ~ & Settings & $\delta$1 $\uparrow$ & $\delta$2 $\uparrow$ & $\delta$3 $\uparrow$ & Abs Rel $\downarrow$ & RMSE $\downarrow$ & RMSE log $\downarrow$ & FLOPs (G) & Pram (M) \\ 
\midrule
        \multirow{7}{*}{Components} & W/o EIFusion & 0.877 & 0.978 & 0.995 & 0.110 & 3.885 & 0.148 & 80 & 57.6 \\
        ~ & W/o CA in decoder & 0.869 & 0.977 & \textbf{0.996} & 0.111 & 4.074 & 0.153 & 80 & 57.3  \\ 
        ~ & W/o GRU & 0.858 & 0.978 & 0.995 & 0.115 & 4.234 & 0.158 & 69 & 56.8 \\ 
        ~ & W/o context in decoder & 0.876 & 0.977 & 0.995 & 0.112 & 3.921 & 0.150 & 73 & 57.1 \\ 
        ~ & W/o multi-scale in lookup & 0.875 & 0.977 & 0.995 & 0.110 & 3.969 & 0.151 & 76 & 57.4 \\ 
        ~ & W/o DHV & 0.873 & 0.977 & 0.995 & 0.113 & 3.964 & 0.152 & 81 & 57.4 \\ 
        ~ & \underline{Ours} & \textbf{0.885} & \textbf{0.979} & \textbf{0.996} & \textbf{0.105} & \textbf{3.857} & \textbf{0.145} & 80 & 57.4 \\ 
\midrule
        \multirow{4}{*}{Iteration} & 2 & 0.869 & 0.977 & 0.995 & 0.117 & 3.930 & 0.152 & 50 & 57.4 \\
        ~ & 4 & 0.876 & 0.978 & 0.995 & 0.111 & 3.878 & 0.149 & 60 & 57.4  \\ 
        ~ & \underline{8} & \textbf{0.885} & \textbf{0.979} & \textbf{0.996} & \textbf{0.105} & \textbf{3.857} & \textbf{0.145} & 80 & 57.4 \\ 
        ~ & 12 & 0.862 & 0.975 & 0.994 & 0.117 & 4.183 & 0.159 & 100 & 57.4 \\
\midrule 
        \multirow{4}{*}{Radius} & 0 & 0.868 & 0.974 & 0.993 & 0.116 & 4.052 & 0.155 & 74 & 57.3 \\
        ~ & 2 & 0.877 & 0.978 & 0.995 & 0.111 & 3.916 & 0.150 & 75 & 57.4 \\ 
        ~ & 4 & 0.881 & \textbf{0.979} & \textbf{0.996} & 0.108 & 3.858 & 0.147 & 77 & 57.4 \\ 
        ~ & \underline{8} & \textbf{0.885} & \textbf{0.979} & \textbf{0.996} & \textbf{0.105} & \textbf{3.857} & \textbf{0.145} & 80 & 57.4 \\ 
\midrule 
        \multirow{3}{*}{Number of $D$} & 32, 32, 32 & 0.870 & 0.975 & 0.994 & 0.114 & 4.010 & 0.155 & 80 & 57.4 \\
        ~ & 32, 64, 64 & 0.877 & 0.978 & 0.995 & 0.111 & 3.947 & 0.151 & 80 & 57.4 \\ 
        ~ & \underline{32, 64, 96} & \textbf{0.885} & \textbf{0.979} & \textbf{0.996} & \textbf{0.105} & \textbf{3.857} & \textbf{0.145} & 80 & 57.4 \\ 
\midrule 
        \multirow{3}{*}{Resolution} & 1/16 & 0.834 & 0.967 & 0.993 & 0.126 & 4.378 & 0.169 & 46  & 56.6 \\
        ~ & 1/16, 1/8 & 0.860  & 0.975 & 0.994  & 0.119  & 4.068  & 0.157 & 57 & 57.1 \\ 
        ~ & \underline{1/16, 1/8, 1/4} & \textbf{0.885} & \textbf{0.979} & \textbf{0.996} & \textbf{0.105} & \textbf{3.857} & \textbf{0.145} & 80 & 57.4 \\ 
\bottomrule 
    \end{tabular}}
\end{table*}

\begin{figure}[t]
\centering
\includegraphics[width=0.32\textwidth,height=0.22\textwidth]{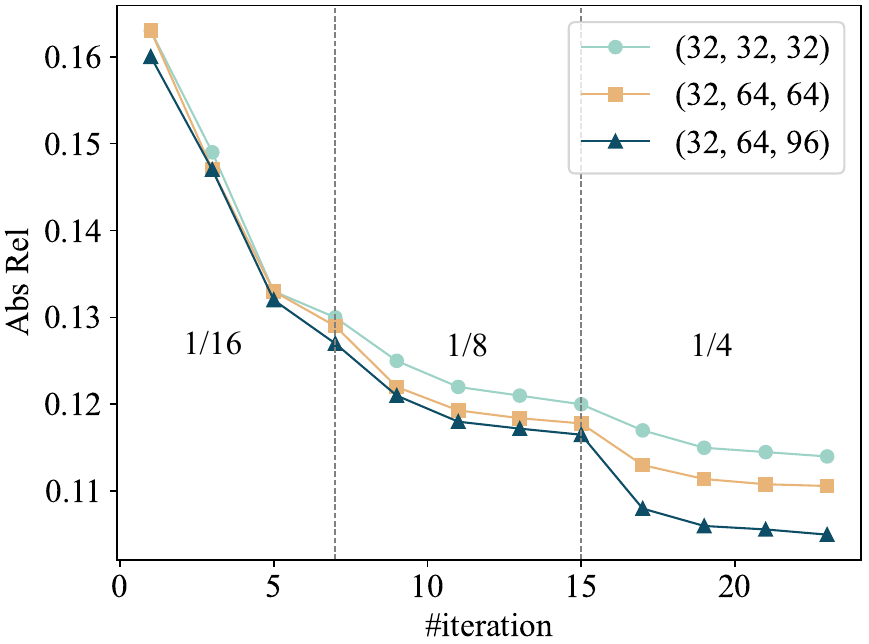}
\caption{The refinement of Abs Rel under different $D$ settings.}
\label{dvals}
\end{figure}

\begin{figure*}[t]
\centering
\includegraphics[width=0.80\textwidth,height=0.16\textwidth]{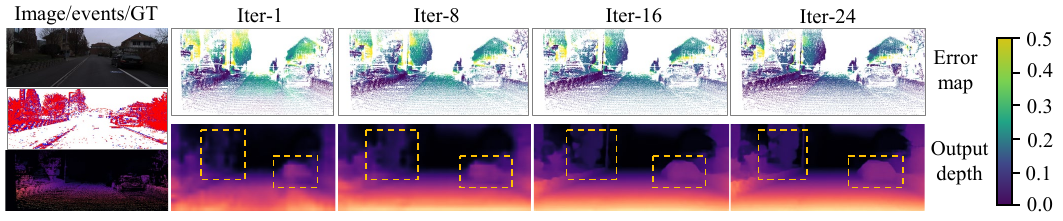}
\caption{Visualization of multi-resolution iterative depth refinement and the evolution of error maps across iterations.}
\label{multilevel}
\end{figure*}

\noindent\textbf{W/o GRU.} 
When the GRU unit is replaced with convolution layers, the absence of hidden states prevents the model from propagating information from previous iterations, leading to weaker convergence due to the lack of prior guidance. 

\noindent\textbf{W/o Context in Decoder.} 
We attempt to decode depth without contextual features, relying solely on correlation features. The results show that although correlation features provide cues for optimal depth matching, the absence of scene-level contextual information needed for spatial consistency leads to degraded performance.

\noindent\textbf{W/o Multi-scale in Lookup.} 
Moreover, removing multi-scale pooling in the cost volume reduces robustness across different scenarios, confirming the importance of multi-scale lookup for reliable depth estimation.

\noindent\textbf{W/o DHV.} 
We replace the discrete DHV with the continuous depth from the previous iteration. For each contextual feature pixel, the cost volume is computed by sampling features within an $r$-neighborhood in the depth embedding map. However, the DHV-based formulation performs better, as coarse depth predictions accumulate errors over iterations, whereas depth search within the DHV offers error tolerance, and the geometry encoder enhances consistency.

\noindent\textbf{Iterations.} 
The results in~\cref{ablation} show that increasing the iteration gradually improves depth estimation accuracy. However, more iterations increase computations and may destabilize training, preventing the model from converging to the optimal solution.

\noindent\textbf{Lookup Radius.} 
The results show that a larger lookup radius provides richer matching cues and improves performance. Even when $r=0$, where only single-point matching is available, our model outperforms recent works~\cite{EReformer, depthanyevent}, confirming its effectiveness.

\noindent\textbf{Number of $D$ in DHV.} 
We investigate the impact of the number of depth values ($D$). Increasing $D$ improves performance without increasing computations. A larger $D$ at higher resolutions provides finer DHV, allowing the model to perform more precise matching and optimization. Therefore, using DHVs with different granularities across feature levels accelerates convergence and yields more stable and accurate depth estimation, as shown in~\cref{dvals}.

\noindent\textbf{Multi-resolution Refinement.}
Our lightweight 3D cost volume allows depth refinement at high resolutions. The results show that progressively refining from low to high resolutions achieves the best performance. The 1/16 stage focuses on global consistency to produce coarse depth, while the 1/8 and 1/4 stages further refine structural details. As shown in~\cref{multilevel}, the regions with reduced errors (reduction of the yellow part) correspond to more fine-grained and accurate depth estimates.

\section{Conclusion}
\label{sec:Conclusion}
In this work, we propose HypoDepth, a novel event-image-based MDE iterative refinement framework. By introducing a Depth Hypothesis Volume (DHV), we reformulate the ill-posed depth regression problem as a depth search problem. We construct a 3D cost volume between DHV and contextual features for iterative lookup–based residual optimization, effectively reducing the challenges of full-depth regression. Benefits from the lightweight cost volume, we perform refinement across multiple resolutions, achieving robust global-to-fine depth estimation. Our method shows the best performance on DSEC and MVSEC, and demonstrates superior zero-shot generalization.

\noindent\textbf{Acknowledgment.}
This work was supported by the National Natural Science Foundation of China (Grant No. 62273093, 62236002).

{
    \small
    \bibliographystyle{ieeenat_fullname}
    \bibliography{main}

@String(CVPR= {IEEE Conf. Comput. Vis. Pattern Recog.})

@String(ICCV= {Int. Conf. Comput. Vis.})

@String(ECCV= {Eur. Conf. Comput. Vis.})

@String(CVPR  = {CVPR})

@String(ICCV  = {ICCV})

@String(ECCV  = {ECCV})

@InProceedings{rayfusion,
author={Kim, Jungeon and Kim, Soongjin and Park, Jaesik and Lee, Seungyong},
title={Deep Cost Ray Fusion for Sparse Depth Video Completion},
booktitle={European Conference on Computer Vision (ECCV)},
year={2025},
pages={328--346},
isbn={978-3-031-73347-5}
}

@ARTICLE{dvs,
  author={Lichtsteiner, Patrick and Posch, Christoph and Delbruck, Tobi},
  journal={IEEE Journal of Solid-State Circuits}, 
  title={A 128$\times$ 128 120 dB 15 $\mu$s Latency Asynchronous Temporal Contrast Vision Sensor}, 
  year={2008},
  volume={43},
  number={2},
  pages={566-576},
  doi={10.1109/JSSC.2007.914337}
}

@ARTICLE{mde_residuals,
  author={Song, Minsoo and Lim, Seokjae and Kim, Wonjun},
  journal={IEEE Transactions on Circuits and Systems for Video Technology}, 
  title={Monocular Depth Estimation Using Laplacian Pyramid-Based Depth Residuals}, 
  year={2021},
  volume={31},
  number={11},
  pages={4381-4393},
  doi={10.1109/TCSVT.2021.3049869}
}

@Inproceedings{DPT,
  title={Vision transformers for dense prediction},
  author={Ranftl, Ren{\'e} and Bochkovskiy, Alexey and Koltun, Vladlen},
  booktitle={Proceedings of the IEEE/CVF International Conference on Computer Vision (ICCV)},
  pages={12179--12188},
  year={2021}
}

@article{zhang2022hierarchical,
  title={Hierarchical normalization for robust monocular depth estimation},
  author={Zhang, Chi and Yin, Wei and Wang, Billzb and Yu, Gang and Fu, Bin and Shen, Chunhua},
  journal={Advances in Neural Information Processing Systems (NeurIPS)},
  volume={35},
  pages={14128--14139},
  year={2022}
}

@article{ranftl2020towards,
  title={Towards robust monocular depth estimation: Mixing datasets for zero-shot cross-dataset transfer},
  author={Ranftl, Ren{\'e} and Lasinger, Katrin and Hafner, David and Schindler, Konrad and Koltun, Vladlen},
  journal={IEEE Transactions on Pattern Analysis and Machine Intelligence},
  volume={44},
  number={3},
  pages={1623--1637},
  year={2020},
}

@ARTICLE{survey,
  author={Gallego, Guillermo and Delbrück, Tobi and Orchard, Garrick and Bartolozzi, Chiara and Taba, Brian and Censi, Andrea and Leutenegger, Stefan and Davison, Andrew J. and Conradt, Jörg and Daniilidis, Kostas and Scaramuzza, Davide},
  journal={IEEE Transactions on Pattern Analysis and Machine Intelligence}, 
  title={Event-Based Vision: A Survey}, 
  year={2022},
  volume={44},
  number={1},
  pages={154-180}
}

@article{RAMNet,
  title={Combining events and frames using recurrent asynchronous multimodal networks for monocular depth prediction},
  author={Gehrig, Daniel and R{\"u}egg, Michelle and Gehrig, Mathias and Hidalgo-Carri{\'o}, Javier and Scaramuzza, Davide},
  journal={IEEE Robotics and Automation Letters},
  volume={6},
  number={2},
  pages={2822--2829},
  year={2021},
}

@Inproceedings{depthanyevent,
  title={Depth AnyEvent: A Cross-Modal Distillation Paradigm for Event-Based Monocular Depth Estimation},
  author={Bartolomei, Luca and Mannocci, Enrico and Tosi, Fabio and Poggi, Matteo and Mattoccia, Stefano},
  booktitle={Proceedings of the IEEE/CVF International Conference on Computer Vision (ICCV)},
  pages={19669--19678},
  year={2025}
}

@article{EReformer,
  title={Event-based monocular dense depth estimation with recurrent transformers},
  author={Liu, Xu and Li, Jianing and Fan, Xiaopeng and Tian, Yonghong},
  journal={arXiv preprint arXiv:2212.02791},
  year={2022}
}

@article{fuse,
  title={FUSE: Label-Free Image-Event Joint Monocular Depth Estimation via Frequency-Decoupled Alignment and Degradation-Robust Fusion},
  author={Sun, Pihai and Jiang, Junjun and Yao, Yuanqi and Chen, Youyu and Zhao, Wenbo and Jiang, Kui and Liu, Xianming},
  journal={arXiv preprint arXiv:2503.19739},
  year={2025}
}

@Inproceedings{HMNet,
  title={Hierarchical neural memory network for low latency event processing},
  author={Hamaguchi, Ryuhei and Furukawa, Yasutaka and Onishi, Masaki and Sakurada, Ken},
  booktitle={Proceedings of the IEEE/CVF Conference on Computer Vision and Pattern Recognition (CVPR)},
  pages={22867--22876},
  year={2023}
}

@inproceedings{e2depth,
  title={Learning monocular dense depth from events},
  author={Hidalgo-Carri{\'o}, Javier and Gehrig, Daniel and Scaramuzza, Davide},
  booktitle={International Conference on 3D Vision (3DV)},
  pages={534--542},
  year={2020},
}

@Inproceedings{pcdepth,
  title={PCDepth: Pattern-based complementary learning for monocular depth estimation by best of both worlds},
  author={Liu, Haotian and Qu, Sanqing and Lu, Fan and Bu, Zongtao and Roehrbein, Florian and Knoll, Alois and Chen, Guang},
  booktitle={IEEE/RSJ International Conference on Intelligent Robots and Systems (IROS)},
  pages={11187--11194},
  year={2024},
}

@inproceedings{zhu2023self,
  title={Self-supervised event-based monocular depth estimation using cross-modal consistency},
  author={Zhu, Junyu and Liu, Lina and Jiang, Bofeng and Wen, Feng and Zhang, Hongbo and Li, Wanlong and Liu, Yong},
  booktitle={IEEE/RSJ International Conference on Intelligent Robots and Systems (IROS)},
  pages={7704--7710},
  year={2023},
}

@Inproceedings{srfnet,
  title={Srfnet: Monocular depth estimation with fine-grained structure via spatial reliability-oriented fusion of frames and events},
  author={Pan, Tianbo and Cao, Zidong and Wang, Lin},
  booktitle={IEEE International Conference on Robotics and Automation (ICRA)},
  pages={10695--10702},
  year={2024},
}

@Inproceedings{tespec,
  title={TESPEC: Temporally-Enhanced Self-Supervised Pretraining for Event Cameras},
  author={Mohammadi, Mohammad and Wu, Ziyi and Gilitschenski, Igor},
  booktitle={Proceedings of the IEEE/CVF International Conference on Computer Vision (ICCV)},
  pages={7782--7793},
  year={2025}
}

@article{unict,
  title={UniCT Depth: Event-Image Fusion Based Monocular Depth Estimation with Convolution-Compensated ViT Dual SA Block},
  author={Jing, Luoxi and Shi, Dianxi and Liu, Zhe and Jin, Songchang and Qiu, Chunping and Qiao, Ziteng and Li, Yuxian and Xia, Jianqiang},
  journal={arXiv preprint arXiv:2507.19948},
  year={2025}
}

@Inproceedings{raft,
  title={Raft: Recurrent all-pairs field transforms for optical flow},
  author={Teed, Zachary and Deng, Jia},
  booktitle={European Conference on Computer Vision (ECCV)},
  pages={402--419},
  year={2020},
}

@inproceedings{laina2016deeper,
  title={Deeper depth prediction with fully convolutional residual networks},
  author={Laina, Iro and Rupprecht, Christian and Belagiannis, Vasileios and Tombari, Federico and Navab, Nassir},
  booktitle={International Conference on 3D Vision (3DV)},
  pages={239--248},
  year={2016},
}

@article{eigen2014depth,
  title={Depth map prediction from a single image using a multi-scale deep network},
  author={Eigen, David and Puhrsch, Christian and Fergus, Rob},
  journal={Advances in Neural Information Processing Systems (NeurIPS)},
  volume={27},
  year={2014}
}

@inproceedings{bhat2021adabins,
  title={Adabins: Depth estimation using adaptive bins},
  author={Bhat, Shariq Farooq and Alhashim, Ibraheem and Wonka, Peter},
  booktitle={Proceedings of the IEEE/CVF Conference on Computer Vision and Pattern Recognition (CVPR)},
  pages={4009--4018},
  year={2021}
}

@inproceedings{fu2018deep,
  title={Deep ordinal regression network for monocular depth estimation},
  author={Fu, Huan and Gong, Mingming and Wang, Chaohui and Batmanghelich, Kayhan and Tao, Dacheng},
  booktitle={Proceedings of the IEEE/CVF Conference on Computer Vision and Pattern Recognition (CVPR)},
  pages={2002--2011},
  year={2018}
}

@inproceedings{yang2024depth,
  title={Depth anything: Unleashing the power of large-scale unlabeled data},
  author={Yang, Lihe and Kang, Bingyi and Huang, Zilong and Xu, Xiaogang and Feng, Jiashi and Zhao, Hengshuang},
  booktitle={Proceedings of the IEEE/CVF Conference on Computer Vision and Pattern Recognition (CVPR)},
  pages={10371--10381},
  year={2024}
}

@article{yang2024depth2,
  title={Depth anything v2},
  author={Yang, Lihe and Kang, Bingyi and Huang, Zilong and Zhao, Zhen and Xu, Xiaogang and Feng, Jiashi and Zhao, Hengshuang},
  journal={Advances in Neural Information Processing Systems (NeurIPS)},
  volume={37},
  pages={21875--21911},
  year={2024}
}

@article{BTS,
  title={From big to small: Multi-scale local planar guidance for monocular depth estimation},
  author={Lee, Jin Han and Han, Myung-Kyu and Ko, Dong Wook and Suh, Il Hong},
  journal={arXiv preprint arXiv:1907.10326},
  year={2019}
}

@inproceedings{duan2024diffusiondepth,
  title={Diffusiondepth: Diffusion denoising approach for monocular depth estimation},
  author={Duan, Yiquan and Guo, Xianda and Zhu, Zheng},
  booktitle={European Conference on Computer Vision (ECCV)},
  pages={432--449},
  year={2024},
}

@article{shao2024monodiffusion,
  title={Monodiffusion: Self-supervised monocular depth estimation using diffusion model},
  author={Shao, Shuwei and Pei, Zhongcai and Chen, Weihai and Sun, Dingchi and Chen, Peter CY and Li, Zhengguo},
  journal={IEEE Transactions on Circuits and Systems for Video Technology},
  year={2024},
}

@inproceedings{raftstereo,
  title={Raft-stereo: Multilevel recurrent field transforms for stereo matching},
  author={Lipson, Lahav and Teed, Zachary and Deng, Jia},
  booktitle={International Conference on 3D Vision (3DV)},
  pages={218--227},
  year={2021},
}

@Inproceedings{TMA,
  title={TMA: Temporal motion aggregation for event-based optical flow},
  author={Liu, Haotian and Chen, Guang and Qu, Sanqing and Zhang, Yanping and Li, Zhijun and Knoll, Alois and Jiang, Changjun},
  booktitle={Proceedings of the IEEE/CVF International Conference on Computer Vision (ICCV)},
  pages={9685--9694},
  year={2023}
}

@Inproceedings{edcflow,
  title={EDCFlow: Exploring Temporally Dense Difference Maps for Event-based Optical Flow Estimation},
  author={Liu, Daikun and Cheng, Lei and Wang, Teng and Sun, Changyin},
  booktitle={Proceedings of the IEEE/CVF Conference on Computer Vision and Pattern Recognition (CVPR)},
  pages={1984--1993},
  year={2025}
}

@Inproceedings{eraft,
  title={E-raft: Dense optical flow from event cameras},
  author={Gehrig, Mathias and Millh{\"a}usler, Mario and Gehrig, Daniel and Scaramuzza, Davide},
  booktitle={International Conference on 3D Vision (3DV)},
  pages={197--206},
  year={2021},
}

@inproceedings{zhu_unflow,
  title={Unsupervised event-based learning of optical flow, depth, and egomotion},
  author={Zhu, Alex Zihao and Yuan, Liangzhe and Chaney, Kenneth and Daniilidis, Kostas},
  booktitle={Proceedings of the IEEE/CVF Conference on Computer Vision and Pattern Recognition (CVPR)},
  pages={989--997},
  year={2019}
}

@article{xu2025igev++,
  title={Igev++: Iterative multi-range geometry encoding volumes for stereo matching},
  author={Xu, Gangwei and Wang, Xianqi and Zhang, Zhaoxing and Cheng, Junda and Liao, Chunyuan and Yang, Xin},
  journal={IEEE Transactions on Pattern Analysis and Machine Intelligence},
  year={2025},
}

@article{dsec,
  title={Dsec: A stereo event camera dataset for driving scenarios},
  author={Gehrig, Mathias and Aarents, Willem and Gehrig, Daniel and Scaramuzza, Davide},
  journal={IEEE Robotics and Automation Letters},
  volume={6},
  number={3},
  pages={4947--4954},
  year={2021},
}

@article{mvsec,
  title={The multivehicle stereo event camera dataset: An event camera dataset for 3D perception},
  author={Zhu, Alex Zihao and Thakur, Dinesh and {\"O}zaslan, Tolga and Pfrommer, Bernd and Kumar, Vijay and Daniilidis, Kostas},
  journal={IEEE Robotics and Automation Letters},
  volume={3},
  number={3},
  pages={2032--2039},
  year={2018},
}

@inproceedings{liu2021swin,
  title={Swin transformer: Hierarchical vision transformer using shifted windows},
  author={Liu, Ze and Lin, Yutong and Cao, Yue and Hu, Han and Wei, Yixuan and Zhang, Zheng and Lin, Stephen and Guo, Baining},
  booktitle={Proceedings of the IEEE/CVF International Conference on Computer Vision (ICCV)},
  pages={10012--10022},
  year={2021}
}

@inproceedings{bartolomei2024lidar,
  title={Lidar-event stereo fusion with hallucinations},
  author={Bartolomei, Luca and Poggi, Matteo and Conti, Andrea and Mattoccia, Stefano},
  booktitle={European Conference on Computer Vision (ECCV)},
  pages={125--145},
  year={2024},
}

@article{adamw,
  title={Decoupled weight decay regularization},
  author={Loshchilov, I},
  journal={arXiv preprint arXiv:1711.05101},
  year={2017}
}

@inproceedings{onecycle,
  title={Super-convergence: Very fast training of neural networks using large learning rates},
  author={Smith, Leslie N and Topin, Nicholay},
  booktitle={Artificial Intelligence and Machine Learning for Multi-domain Operations Applications},
  volume={11006},
  pages={369--386},
  year={2019},
}

@article{liu2025high,
  title={High-Rate Monocular Depth Estimation via Cross Frame-Rate Collaboration of Frames and Events},
  author={Liu, Xu and Fan, Xiaopeng and Li, Jianing and Li, Dianze and Zhang, Wei and Ma, Zhengyu and Tian, Yonghong},
  journal={International Journal of Computer Vision},
  volume={133},
  number={10},
  pages={7332--7351},
  year={2025},
  publisher={Springer}
}

@InProceedings{Li_2025_CVPR,
    author    = {Li, Jianing and Zhang, Yunjian and Han, Haiqian and Ji, Xiangyang},
    title     = {Active Event-based Stereo Vision},
    booktitle = {Proceedings of the IEEE/CVF Conference on Computer Vision and Pattern Recognition (CVPR)},
    month     = {June},
    year      = {2025},
    pages     = {971-981}
}
}

\clearpage
\setcounter{page}{1}
\maketitlesupplementary


\section{Network Details}
\noindent\textbf{HypoDepth.}
Our model performs depth refinement progressively from 1/16 to 1/4 resolution. The detailed network configurations are summarized in~\cref{network_settings}, including the backbone, channel dimensions of core modules, the number of iterative updates, and the parameters related to cost-volume lookup. \textbf{Ours} denotes the event–image fusion model, while \textbf{Ours-E} represents the event-only variant, in which the image encoder and EIFusion module are removed. Notably, we adopt resolution-dependent lookup scales ($l$ in~\cref{lookup} (a)) across different stages in our base model. At lower resolutions (such as 1/16 resolution), the lookup radius $r$ already covers most or even all spatial regions of the feature map, making the resulting correlation effectively global. Further downsampling would cause the lookup region to span nearly the entire feature map again, leading to redundant correlations and reducing the effectiveness of the multi-scale lookup strategy. For the scaled-down model in the ablation studies, we train it with a batch size of 3 for 40k iterations.

\noindent\textbf{PCDepth-Ours.}
Since PCDepth~\cite{pcdepth} performs iterative feature-level fusion refinement at 1/8 resolution for depth optimization, we also conduct refinement at the same scale for fair comparison. The contextual features are obtained from PCDepth’s fused encoder output, while the parameters of our iterative refinement decoder are listed in~\cref{network_settings}.

\section{More Results}
\begin{figure}[t]
\centering
  \includegraphics[width=0.5\linewidth, height = 0.3\linewidth]{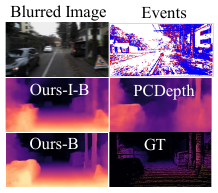}
  \caption{Qualitative results on blurred DSEC dataset.}
   \label{motionblur}
\end{figure}

\noindent\textbf{More Comparison on MVSEC and DSEC.}
In~\cref{mvsec_blurreddsec}, we report the results of our event-only (Ours-E-B) and image-only (Ours-I-B) variants on DSEC and MVSEC to verify the benefit of event–image fusion. In addition, we simulate image motion blur on DSEC to mimic high-speed scenarios, and directly evaluate the model trained on the original DSEC under zero-shot transfer to the blurred setting.~\cref{mvsec_blurreddsec} compares our image-only variant (w/o events) with PCDepth in this challenging scenario. The results show that incorporating events consistently improves performance, with larger gains in challenging conditions such as nighttime (MVSEC\_on1) and high-speed motion (blurred DSEC). In these cases, frame appearance is severely degraded by low illumination or motion blur, whereas events still provide rich structural and texture cues, leading to more reliable correspondence and estimation; in contrast, frame-only methods degrade notably. Moreover, our method generalizes better than PCDepth under high-speed blur. The qualitative comparisons in~\cref{motionblur} further confirm our advantage in the presence of image blur.

\begin{table}[t]
\centering
\caption{More results on MVSEC~\cite{mvsec}, DSEC~\cite{dsec}, and blurred DSEC. The best are in bold.} 
\label{mvsec_blurreddsec} 
    \resizebox{1.0\linewidth}{!}{
\begin{tabular}{cccccc}
    \toprule
            Dataset & Input & Method & $\delta$1 $\uparrow$ & Abs Rel $\downarrow$ & RMSE $\downarrow$ \\ 
    \midrule 
             \multirow{3}{*}{DSEC} & I & Ours-I-B & 0.859 & 0.116 & 4.183 \\
                ~ & E & Ours-E-B & 0.839 & 0.129 & 4.442 \\
                ~ & E+I & \textbf{Ours-B} & \textbf{0.901} & \textbf{0.099} & \textbf{3.583}\\
            \midrule 
                \multirow{4}{*}{MVSEC\_od1 / on1} & I & Ours-I-B & 0.692 / 0.534 & 0.233 / 0.330 & 7.020/8.537 \\
                ~ & E & Ours-E-B & 0.690 / 0.600 & 0.243 / 0.289 & 6.732 / 7.201 \\
                                ~ & E+I & Ours-T & 0.682 / 0.573 & 0.260 / 0.312 & 7.157 / 7.597 \\
                ~ & E+I & \textbf{Ours-B} & \textbf{0.698} / \textbf{0.614} & \textbf{0.212} / \textbf{0.268} & \textbf{6.556} / \textbf{6.834}\\
    \midrule 
                \multirow{3}{*}{Blurred DSEC}  & I & Ours-I-B  & 0.121 & 0.5207 & 15.9315 \\
            ~ & E+I & \textbf{Ours-B}   & \textbf{0.665} & \textbf{0.1828} & \textbf{6.3642} \\
            ~ & E+I & PCDepth  & 0.209 & 0.454 & 14.627 \\
            \bottomrule 
    \end{tabular}}
\end{table}

\noindent\textbf{Duration of Event Slices.}
We investigate the impact of different event-stream window lengths, including 20 ms, 50 ms, and 100 ms. Results in \cref{duration} show that 20 ms and 50 ms achieve comparable performance, indicating robustness to the temporal window. In contrast, longer windows cause excessive event accumulation under large motion, increasing misalignment and motion artifacts, blurring the matching distribution, and degrading performance. We therefore adopt a 20 ms window to reduce latency, better leverage the strengths of event cameras, and ensure a fair performance comparison.

\begin{table}[h]
\centering
\caption{Ablation studies on the duration of event slices.} 
\label{duration} 
    \resizebox{0.7\linewidth}{!}{
    \begin{tabular}{cccc}
\toprule
        Duration (ms) & $\delta$1 $\uparrow$ & Abs Rel $\downarrow$ & RMSE $\downarrow$ \\ 
\midrule 
        \underline{20}  & \underline{0.885} & \textbf{0.105} & \underline{3.857} \\
        50 & \textbf{0.887}& \underline{0.106} & \textbf{3.705} \\
        100 & 0.876& 0.112 & 3.869 \\
\bottomrule 
    \end{tabular}}
\end{table}

\noindent\textbf{Depth-range of DHV.}
We investigate the effect of the depth-range on accuracy. DHV defines a coarse search space, while the subsequent continuous residual updates refine depth and alleviate discretization errors. Experiments show that an overly narrow depth range degrades accuracy, whereas expanding the range further has a marginal impact.

\begin{table}[h]
\centering
\caption{Ablation studies on the depth-range of DHV.}
\label{D-range} 
    \resizebox{0.8\linewidth}{!}{
    \begin{tabular}{cccc}
\toprule
        Depth range & $\delta$1 $\uparrow$ & Abs Rel $\downarrow$ & RMSE $\downarrow$ \\ 
\midrule 
        $[0.05*d_{min}, 2*d_{max}]$ & 0.884 & 0.107 & 3.833  \\
\midrule 
         \underline{$[d_{min}, d_{max}]$} & 0.885 & 0.105 & 3.857 \\
\midrule
        $[2*d_{min}, 0.7*d_{max}]$ & 0.793 & 0.141 & 4.533 \\
\bottomrule 
    \end{tabular}}
\end{table}


\section{Visualizations}
\begin{figure*}[t]
\centering
\includegraphics[width=1.0\textwidth,height=0.35\textwidth]{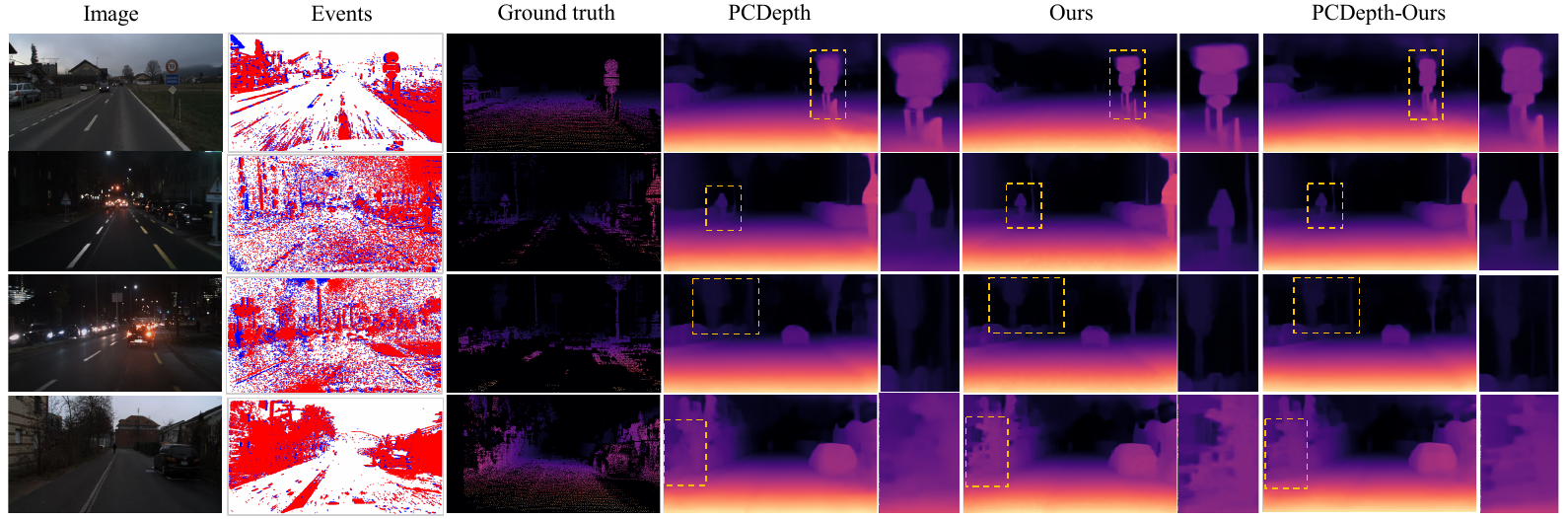}
\caption{More qualitative monocular depth estimation results on DSEC~\cite{dsec}. The first and fourth rows correspond to daytime scenes, while the second and third rows represent night scenes.}
\label{more_dsec_viz}
\end{figure*}

\noindent\textbf{Qualitative Results on DSEC.}
In~\cref{more_dsec_viz}, we present additional qualitative comparisons on the DSEC~\cite{dsec} among \textbf{Ours-B}, PCDepth~\cite{pcdepth}, and \textbf{PCDepth-Ours}. The results demonstrate that on this complex-structured dataset, our approach consistently produces more geometrically plausible and spatially accurate depth maps under varying illumination conditions compared to PCDepth. Moreover, when equipped with our iterative refinement module, \textbf{PCDepth-Ours} also benefits from sharper and more accurate object boundaries. Thanks to the additional 1/4-resolution refinement stage in our model, our depth predictions exhibit finer structural details than those of \textbf{PCDepth-Ours}.

\begin{figure*}[t]
\centering
\includegraphics[width=1.0\textwidth,height=0.45\textwidth]{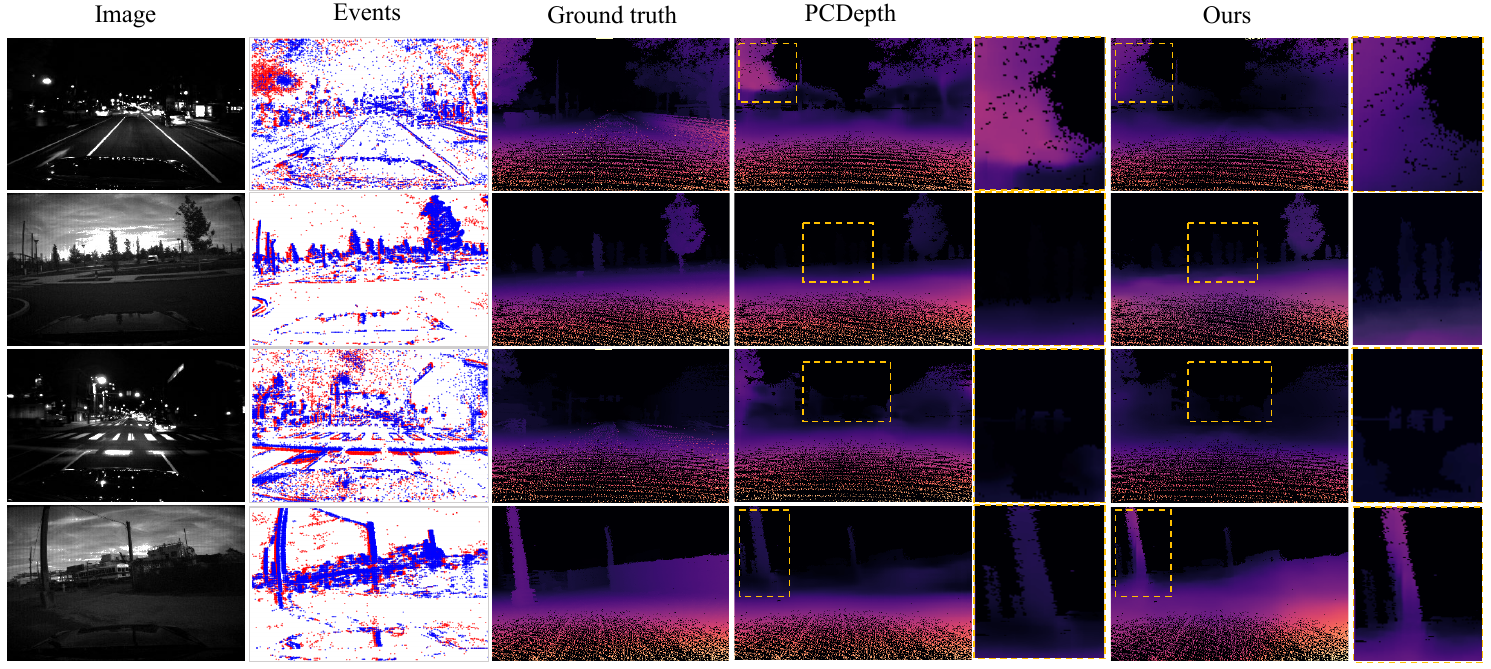}
\caption{Qualitative monocular depth estimation results on MVSEC~\cite{mvsec}. The first and third rows correspond to night scenes, while the second and fourth rows represent daytime scenes.}
\label{mvsec_viz}
\end{figure*}

\noindent\textbf{Qualitative Results on MVSEC.}
We present a qualitative comparison between our method and PCDepth~\cite{pcdepth} on both daytime and nighttime sequences of the MVSEC~\cite{mvsec} in~\cref{mvsec_viz}. The visual results demonstrate that our approach accurately estimates both distant (rows 2 and 3) and nearby (rows 1 and 4) structures under challenging lighting conditions and sparse event inputs. In particular, PCDepth fails to recover distant regions, whereas our model produces stable and continuous estimates, which are crucial for enabling autonomous systems to perceive distant scenes in advance.

\begin{figure*}[t]
\centering
\includegraphics[width=1.0\textwidth,height=0.55\textwidth]{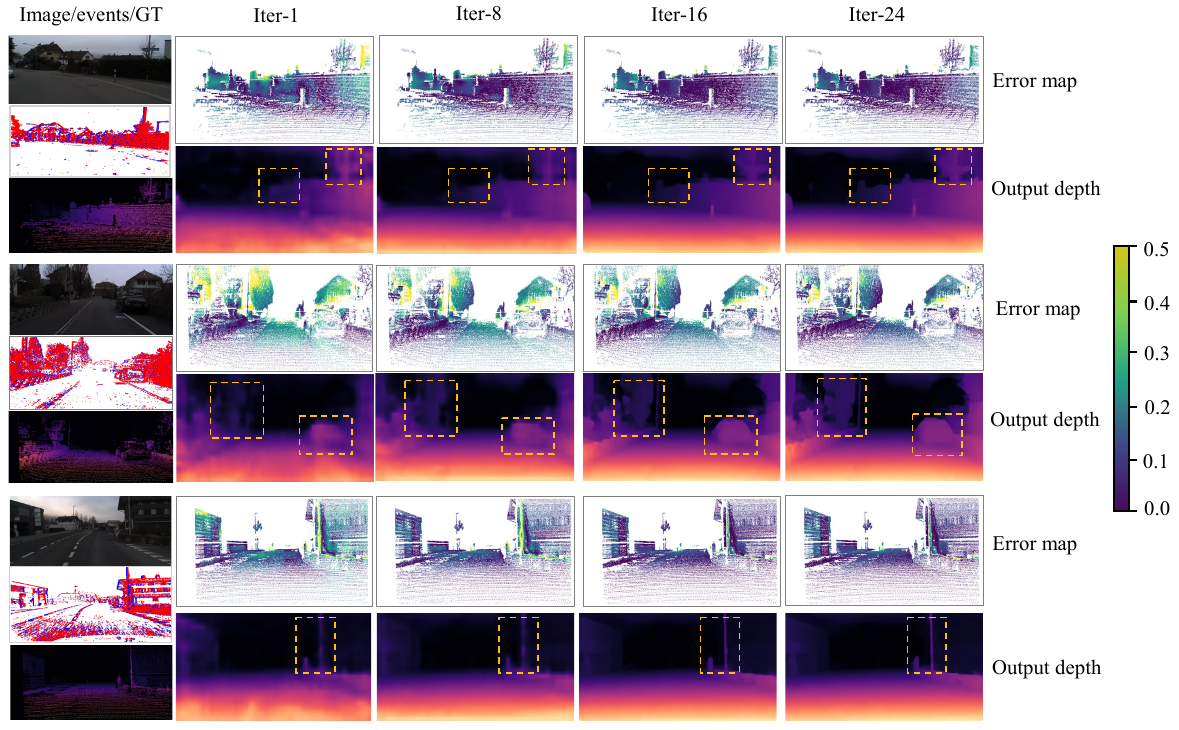}
\caption{Qualitative results on the outdoor\_day1 sequence on MVSEC~\cite{mvsec}.}
\label{more_error}
\end{figure*}

\noindent\textbf{More visualizations of refinement on DSEC.}
We visualize additional multi-resolution depth refinement processes and corresponding error in~\cref{more_error}. At the 1/16 resolution, the model performs globally consistent regression, producing coarse estimations. When refined at the 1/8 level, the scene geometry becomes more pronounced. Further refinement at the 1/4 resolution sharpens both distant and nearby structures, demonstrating the effectiveness of our method in achieving coarse-to-fine depth reconstruction.

\section{Limitations and Future Work} 

\begin{figure}[t]
\centering
\includegraphics[width=0.5\textwidth,height=0.4\textwidth]{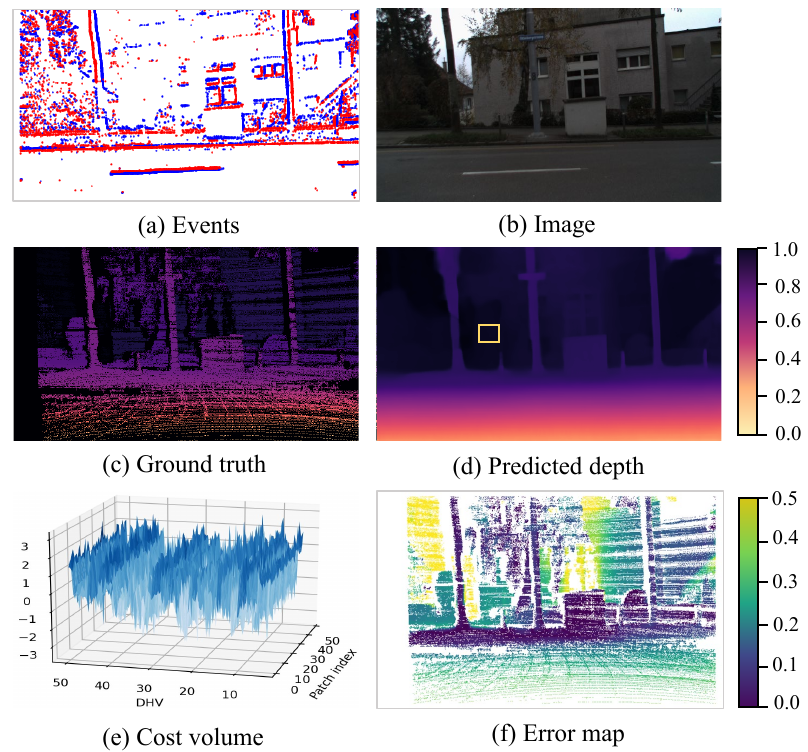}
\caption{Failure case analysis. (e) shows the cost-volume distribution of the yellow patch in (d). (f) shows the log-normalized error. Large errors appear between the predicted depth and the ground truth. In particular, low-texture or homogeneous regions lead to ambiguous matching patterns in the cost volume, resulting in unreliable depth refinement.}
\label{fail_case}
\end{figure}

\begin{table*}[t]
    \centering
    \caption{Detailed network parameter settings. \textbf{PCDepth-Ours} only performs iterative refinement at 1/8 resolution. The values in `()' correspond to the parameters for stages 1/16, 1/8, and 1/4, respectively.} 
    \label{network_settings} 
    \resizebox{1.0\linewidth}{!}{
    \begin{tabular}{ccccccc}
\toprule 
        ~ & Ours-B & Ours-E-B & Ours-E-S & Ours-E-T & PCDepth-Ours & Ours scaled-down \\ 
\midrule
        Backbone & Swin-T & Swin-T & Swin-T & ResNet & Swin-T & Swin-T \\
        Backbone out dims & (128, 96, 64) & (128, 96, 64) & (96, 64, 32) & (64, 64, 32) & - & (128, 64, 48) \\
        EIFusion out dims & (128, 96, 64) & - & - & - & - & (128, 64, 48) \\
        DHV embedding dims & (128, 96, 64) & (128, 96, 64) & (96, 64, 32) & (64, 64, 32) & 128 & (128, 64, 48) \\
        Iterations & (8, 8, 8) & (8, 8, 8) & (3, 3, 3) & (2, 2, 1) & 8 & (8, 8, 8) \\
        $D$ values of DHV & (32, 64, 96) & (32, 64, 96) & (32, 64, 96) & (16, 32, 64) & 64 & (32, 64, 96) \\
        Lookup radius $r$ & (8, 8, 8) & (8, 8, 8) & (8, 8, 8) & (4, 4, 4) & 8 & (8, 8, 8) \\
        Multi-scale ($l$) in Lookup & (1, 2, 3) & (1, 2, 3) & (1, 1, 1) & (1, 1, 1) & 1 & (1, 2, 3) \\
        Geometry encoder out dims & (128, 128, 64) & (128, 128, 64) & (96, 96, 48) & (32, 32, 16) & 64 & (96, 64, 32) \\
        Hidden dims of GRU & (128, 96, 64) & (128, 96, 64) & (64, 32, 24) & (32, 32, 16) & 128 & (64, 32, 24) \\
        Mid-channel of head & (192, 128, 96) & (192, 128, 96) & (96, 64, 64) & (32, 32, 16) & 256 & (128, 96, 64) \\
\bottomrule 
    \end{tabular}}
\end{table*}

We develop a depth-hypothesis–guided iterative refinement framework for event–image monocular depth estimation, achieving high-performance results. To further improve the method, we analyze several remaining limitations. 

First, the superior performance of \textbf{PCDepth-Ours} over \textbf{Ours-B} indicates that our multimodal fusion strategy still has space for improvement. A more adaptive fusion mechanism could better leverage the complementary advantages of the two modalities. For instance, emphasizing image features in static regions where events fail, and focusing on event-based spatiotemporal cues when image appearance becomes unreliable. 

Second, as illustrated in ~\cref{fail_case}, our method fails under certain challenging conditions. The camera motion in this scenario is not forward-facing, resulting in weak parallax between foreground and background objects. Consequently, the spatiotemporal structure captured by events is limited, and the geometric boundaries become ambiguous, $e.g.$, between the window and the transformer box beneath it, or between the tree and the house behind it. Additionally, the scene contains textureless, homogeneous regions such as road surfaces and building facades. These areas exhibit little brightness change and therefore generate few events, while also lacking discriminative texture in the image modality. As a result, the cost volume becomes ambiguous (as shown in~\cref{fail_case} (e)), offering limited guidance for reliable matching. In such cases, the iterative optimization tends to rely more on contextual priors and the depth distribution rather than meaningful correlation evidence obtained from lookup operations.

In future work, we will focus on exploring more effective fusion strategies for images and events to leverage their complementary characteristics fully. To address the limitations of depth estimation under non-forward motion, we will develop multi-view depth estimation models with either multiple temporal continuous frames or multi-perspective inputs. Additionally, we will investigate incorporating additional information (such as semantic segmentation, active structured light~\cite{Li_2025_CVPR}, or LiDAR data) to enhance contextual features and alleviate the ambiguity in cost volume matching, or employ attention mechanisms to enhance matching accuracy. Moreover, cross-frame-rate frameworks~\cite{liu2025high} naturally align with the asynchronous, high-temporal-resolution of events. We plan to extend HypoDepth to asynchronous multimodal fusion, preserving the strength of spatial iterative refinement while effectively aggregating temporal event cues to achieve temporally consistent, low-latency, and high-frequency depth estimation.

\end{document}